%% file: collas_main.tex
\documentclass{article} 
\usepackage{collas2026_conference,times}
\usepackage{easyReview}

\input{math_commands.tex}

\usepackage{hyperref}
\hypersetup{
    colorlinks=true,
    linkcolor=red,
    filecolor=magenta,
    urlcolor=blue,
    citecolor=purple,
    pdftitle={Overleaf Example},
    pdfpagemode=FullScreen,
    }
\usepackage{subcaption}
\usepackage{comment}
\usepackage{tables}
\usepackage{tabularx}
\usepackage{booktabs}
\usepackage{multirow}
\usepackage{lipsum}
\usepackage{algorithm}
\usepackage{algorithmic}
\usepackage{wrapfig}
\usepackage{xcolor}

\def\foo#1\endfoo{}
\newcolumntype{L}{@{}>{\foo}l<{\endfoo}}
\newcolumntype{H}{>{\setbox0=\hbox\bgroup}c<{\egroup}@{}}

\title{GLAM: Efficient Continual Learning at Scale via Grouped LoRA Adapter Merging}

\author{Irene Testa, Luigi Quarantiello, Eric Nuertey Coleman
\\
Department of Computer Science\\
University of Pisa\\
Italy \\
\texttt{\{irene.testa,luigi.quarantiello,eric.coleman\}@phd.unipi.it} \\
\And 
Samrat Mukherjee\\
Centre for Machine Intelligence and Data Science \\
Indian Institute of Technology Bombay \\
India \\
\texttt{samratpisv123@gmail.com}\\
\And 
Julio Hurtado \\
Centre for Applications of Mathematical \& Computing Sciences \\
University of Warwick \\
United Kingdom \\
\texttt{julio.hurtado@warwick.ac.uk} \\
\AND
Vincenzo Lomonaco \\
Department of AI, Data and Decision Sciences
\\
LUISS University \\
Italy \\
\texttt{vlomonaco@luiss.it}
}

\collasfinalcopy 

\begin{document}
\maketitle

\begin{abstract}
The ability to learn continuously over time remains a major challenge for modern machine learning systems, even in the era of Foundation Models.
While the rich representations learned by large pre-trained models can partially mitigate \textit{catastrophic forgetting}, they still struggle to adapt efficiently to evolving data distributions.
A key challenge remains, how to continually add new knowledge to a large pretrained model in a way that is scalable and computationally efficient over long task sequences.
In this work, we introduce GLAM, a simple and effective framework for class-incremental continual learning based on LoRA adapters merging.
For each task, GLAM trains a lightweight low-rank adapter with an importance scalar, incurring minimal computational overhead.
Adapters are then pruned, rescaled, and sequentially grouped to enable structured knowledge reuse and limit parameter growth.
At inference, all groups are combined into a single module, ensuring constant computational cost regardless of the number of tasks.
We evaluate GLAM on vision benchmarks with sequences of up to 50 tasks, significantly extending beyond standard protocols.
GLAM achieves the highest accuracy across the evaluated benchmarks. Compared with the baseline attaining the highest average accuracy across benchmarks, it uses 16--19\% of the trainable parameters and reduces training time by approximately 63--74\%, demonstrating efficient and scalable continual learning.
Our source code is publicly available at \url{https://github.com/atlas-luiss/GLAM}.

\end{abstract}

\section{Introduction}
Continual Learning (CL) aims to build models that can learn incrementally from sequences of tasks while retaining previously learned knowledge, 
addressing the long-standing problem of \textit{catastrophic forgetting}~\citep{
mccloskey1989catastrophic, goodfellow2013empirical}
. The emergence of large pre-trained models has introduced new alternatives to achieve this goal,
which are nonetheless highly costly to retrain or fine-tune for each learning experience, making the development of more efficient approaches essential.

\textbf{Parameter-Efficient Fine-Tuning} (PEFT) methods \citep{han2024parameterefficientfinetuninglargemodels} tackle this issue by adapting only a small subset of the weights of the model or introducing a limited number of trainable parameters, while keeping the base model frozen.
Among PEFT techniques, \textbf{Low-Rank Adaptation} (LoRA) \citep{hu2021} has emerged as a popular choice due to its simplicity and effectiveness.
LoRA, like the other PEFT methods, is primarly designed for static learning scenarios, where the goal is to maximize performance on a single task.
As a result, naively applying LoRA adapters is not beneficial in CL scenarios, as these settings require models to acquire new knowledge over time while preserving previously learned information. 


\begin{figure*}
\centering
\begin{subfigure}[b]{.5\textwidth}
  \centering
  \includegraphics[width=\textwidth]{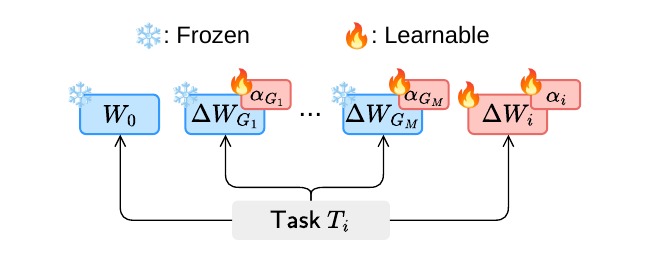} 
  \caption{\footnotesize GLAM first phase: Task-Specific LoRA Training}
  \label{fig:1-phase}
\end{subfigure}
\begin{subfigure}[b]{.49\textwidth}
  \centering
  \includegraphics[width=\textwidth]{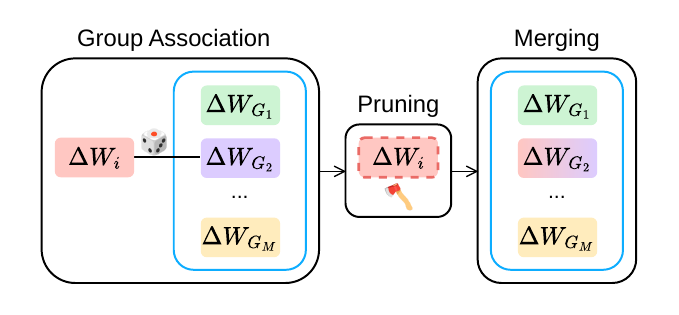}
  \caption{\footnotesize GLAM second phase: Adapters Grouping}
  \label{fig:2-phase}
\end{subfigure}
\caption{Illustration of the GLAM method.
Figure \ref{fig:1-phase}: a new task-specific LoRA adapter $\Delta W_i$ is trained, alongside its importance factor $\alpha_i$ and $\alpha_{G_j}$, one for each of the $M$ groups, where $M$ is much smaller than number of tasks.
Figure \ref{fig:2-phase}: the adapter $\Delta W_i$ is associated with a randomly selected group adapter. After the association, $\Delta W_i$ is pruned and linearly merged with the selected group adapter.
}
\label{fig:method}
\end{figure*}

To adapt PEFT-based approaches to CL scenarios, three key challenges must be addressed: (i) preventing catastrophic forgetting of previously learned tasks, (ii) enabling knowledge transfer between related tasks and (iii) efficiently allocating parameters while maintaining scalability.
Previous LoRA-based approaches \cite{coleman2025parameter} typically address (i) and (ii) by either storing separate adapter modules for each task, requiring task identifiers at inference, or by complex parameter isolation strategies that limit knowledge transfer.
More involved methods \citep{liang2024inflora, wu2025} aim to improve the stability–plasticity trade-off by subspace reparameterization or decoupled magnitude/direction learning.
Nonetheless, they still treat adapters independently throughout training and inference, preventing shared learning or adaptive reuse, and potentially incurring high training costs.
Alternatively, post-hoc merging \citep{marczak2024magmaxleveragingmodelmerging, 10.5555/3666122.3666432, yu2024languagemodelssupermario, ilharco2023editingmodelstaskarithmetic, coleman2024adaptive} consolidates adapters after training, limiting knowledge transfer during the learning phase.
These methods also depend on manually chosen merging coefficients, which can become cumbersome and suboptimal---\textit{especially as the number of tasks grows}.

To address these limitations, we propose \textbf{GLAM}, an efficient LoRA-based continual learning method that dynamically groups and merges adapters as tasks unfold, without requiring any access to previous tasks samples.
For each new task, GLAM
(i) learns a task-specific LoRA adapter along with an associated importance weight,
(ii) prunes and rescales the adapter to reduce potential interference and
(iii) merges it into one of the existing group adapters to limit the parameter growth.
During inference, all group adapters are combined into a single unified module, yielding a model that can be efficiently queried. 
This three-stage framework substantially reduces training overhead, improves computational efficiency, and controls the number of stored modules, while achieving state-of-the-art accuracy and preserving knowledge from prior tasks, thereby mitigating forgetting.
The two combination stages (i.e., within-group and across-group) enhance GLAM’s adaptability and scalability over long task sequences, enabling it to outperform both single-shot merging baselines and more computationally intensive approaches.
We evaluate our method on standard CL benchmarks, with a focus on 
task sequences that are longer than those commonly adopted in recent PEFT-based CL protocols, providing a more demanding \textbf{long-horizon evaluation setting}.
This scenario extends the model’s ``lifetime'' and represents a step toward lifelong learning settings.

To summarize, in this paper we introduce \textbf{GLAM}, a novel LoRA merging approach that addresses the challenge of efficient continual learning over long sequences of tasks through a combination of task-specific adaptations, importance-weighted pruning, and sequential merging.
Our key contributions are as follows:
\begin{itemize}
    \item We propose a scalable CL methodology for foundation models, leveraging PEFT techniques to ensure efficiency, together with a task-specific adaptation mechanism that jointly learns LoRA modules and their importance coefficients;
    \item We design a progressive merging pipeline that consolidates task-specific adapters into a bounded set of group adapters, controlling parameter growth and mitigating merging interference over long task sequences;
    \item We conduct extensive experiments on diverse benchmarks, demonstrating state-of-the-art performance in dynamic CL scenarios over long task sequences while significantly reducing training time and trainable parameters compared to 
    the strongest-performing baselines.
\end{itemize}
Through a comprehensive experimental evaluation, we demonstrate that our method outperforms prior PEFT-based approaches, achieving, on average, a 6\% increase in accuracy and an 8\% reduction in forgetting compared to the strongest baseline.
GLAM is also significantly more efficient than most competing methods, achieving 
approximately a 3$\times$ training speedup over the strongest-performing baseline while using only about 20\% of its trainable parameters.
To gain deeper insights into GLAM's behavior, we conduct extensive ablation studies (Sec. \ref{sec:abl}), which highlight the effectiveness of our sequential, two-phase merging strategy.
These analyses confirm that our approach substantially mitigates task interference, enhances knowledge retention, and preserves scalability and efficiency even over longer task sequences.

\section{Related Work}

\paragraph{Continual Learning}
CL approaches are typically grouped into three categories: (i) regularization-based methods
\citep{kirkpatrick2017overcoming, zenke2017continual}, (ii) replay-based methods
\citep{rebuffi2017icarl,chaudhry2019tiny} and (iii) parameter isolation methods
\citep{mallya2018packnet, rusu2016progressive}.
The rise of large pretrained models has driven the adaptation of CL approaches to Transformer-based architectures \citep{DBLP:journals/corr/VaswaniSPUJGKP17}.
Recent work has increasingly explored how to leverage these models for CL, highlighting both the challenges and opportunities.
On one hand, the rich representations and features extraction abilities captured during pre-training can promote positive forward transfer during sequential learning.
Conversely, the sheer number of parameters complicates the fine-tuning of the model, making it prohibitively expensive and often impractical.

\paragraph{Parameter Efficient Fine-Tuning}
These techniques allow large pre-trained models to be adapted to downstream tasks by modifying only a small set of parameters \citep{houlsby2019parameter}, enabling the learning of new tasks with much higher efficiency than full fine-tuning. Low-Rank Adaptation (LoRA) \citep{hu2021} freezes the pre-trained model weights and injects trainable low-rank decomposition matrices into each layer.
For a pre-trained weight matrix $W_0 \in \mathbb{R}^{d \times k}$, LoRA parameterizes the update $\Delta W$ as the product of two low-rank matrices:

\begin{equation}
    W = W_0 + \Delta W = W_0 + BA
\end{equation}

where $B \in \mathbb{R}^{d \times r}$, $A \in \mathbb{R}^{r \times k}$, and the rank $r \ll \min(d, k)$. This significantly reduces the number of trainable parameters from $d \times k$ to $r \times (d + k)$.
Different LoRA-based techniques have been adapted for CL to enable efficient adaptation over time.
\textit{InfLoRA} \citep{liang2024inflora} proposes an interference-free low-rank adaptation method that reparameterizes pre-trained weights within a subspace designed to minimize interference between tasks.
\textit{SD-LoRA} \cite{wu2025} focuses on dynamic adaptation, which decouples the magnitude and direction of LoRA updates.
\textit{SEMA} \cite{wang2025self} introduces a mixture-of-adapters framework, dynamically expanding the adapters set for new tasks while mitigating forgetting through task-specific routing.  
Nonetheless, such methods fall short when dealing with long task sequences.

An alternative is represented by \textit{prompt-based} methods, which enable continual efficient learning of new tasks through small learnable parameters.
\textit{Learning to Prompt} (L2P) ~\citep{wang2022learning} introduces a framework where a pre-trained model is guided by a set of learnable prompts stored in a memory bank. These prompts are dynamically selected based on input queries, allowing the model to adapt to new tasks without modifying the core parameters.
\textit{DualPrompt}~\citep{wang2022dualprompt} builds upon L2P by incorporating both task-invariant and task-specific prompts. This duality enables the model to capture shared knowledge across tasks while retaining task-specific nuances.
\textit{CODA-Prompt}~\citep{Smith_2023_CVPR} improves prompt-based methods by using an attention-driven key-query mechanism to construct input-conditioned prompts, enhancing adaptability without sacrificing past performance.
However, a key drawback of prompt-based CL methods is their limited plasticity,
which restricts their ability to adapt effectively to novel tasks beyond their pre-training distribution.

\paragraph{Model Merging}
This offers a promising approach to CL by combining expert models to mitigate catastrophic forgetting while leveraging their diverse strengths and capabilities.
In dynamic and ever-evolving domains, this enables models to expand their knowledge while minimizing the loss of prior information.
Different algorithms were defined to perform such combination of multiple models.
Linear merging computes a parameter-wise weighted average, without any particular technique to address interference.
\textit{TIES} \citep{10.5555/3666122.3666432} removes redundant parameters and solves sign conflicts before merging the model's weights.
\textit{DARE} \citep{yu2024languagemodelssupermario} randomly drops a portion of the parameters and rescales the remaining ones to reduce redundancy and minimize interference.

Several works leverage model merging for CL.
\textit{MagMax} \citep{marczak2024magmaxleveragingmodelmerging} fine-tunes tasks sequentially, then consolidates weights via maximum-magnitude selection, requiring no retraining but applying merging only once at the end.
\textit{Orthogonal Projection-Based Continual Merging} \citep{tang2025merging} allows sequential integration of new models, using orthogonal projection to reduce interference and a scaling factor to balance contributions.
\textit{Adaptive LoRA Merging} \citep{coleman2025parameter} replaces fixed-weights combinations with dynamically computed merging coefficients, enhancing task integration.
\textit{MELoRA} \citep{ren2024melora} trains smaller LoRAs in parallel, concatenating them diagonally into a single adapter, increasing representation capacity while maintaining computational efficiency over standard LoRA.

\section{GLAM: Grouped LoRA Adapter Merging}

In this section, we introduce \textbf{GLAM}. We first formalize the problem setting and the objectives of the method (Sec.~\ref{sec:prob_form}). We then describe the GLAM training procedure (Sec.~\ref{sec:glam_training}), which consists of three components: (i) \textbf{task-specific training} of a LoRA adapter for each task, together with its importance coefficient $\alpha$; (ii) \textbf{adapter grouping}, in which newly introduced adapters are assigned to groups and incrementally merged into group-level adapters; and (iii) a \textbf{classifier-alignment stage}, in which the classifier is adapted using synthetic features sampled from stored class statistics. Finally, GLAM consolidates group adapters into a single module through a final adapter concatenation step (Sec.~\ref{sec:final_concat}). Importantly, this final combination can be applied at any point during training, without requiring the entire task sequence to be observed beforehand. An overview of the method is shown in Figure~\ref{fig:method}, while implementation details are reported in Appendix~\ref{app:algorithm}. 

\subsection{Problem Formulation}
\label{sec:prob_form}

We consider the standard class-incremental learning setting, defined by a sequence of tasks
$\mathcal{T} = \{T_1, T_2, \ldots, T_N\}$. At task $T_i$, the learner receives a dataset $D_i = \{(x_j, y_j)\}_{j=1}^{n_i}$ sampled from a distribution $P_i$ over
$\mathcal{X}\times\mathcal{C}_i$, where $x_j\in\mathcal{X}$ is an input sample and
$y_j\in\mathcal{C}_i$ is its label. The input space
$\mathcal{X}\subseteq\mathbb{R}^{3\times H\times W}$ is shared across tasks, while the class sets
$\mathcal{C}_i$ are disjoint. Thus, tasks are defined solely by the class partition.
The model is trained sequentially on the tasks and must retain performance on all previously encountered classes, without access to task identifiers at test time.
We adopt a pre-trained Vision Transformer \citep{Dosovitskiy2020AnII} as the base model and denote its parameters by $W_0$.
The objective is to adapt it to each task in the sequence using low-rank modules, while preventing catastrophic forgetting.
Our goal is to design a strategy that enables effective knowledge transfer between tasks, ultimately making the method particularly computationally efficient and well-suited for longer task sequences.

\paragraph{Notation}
For clarity, we introduce the notation used throughout the method description.
A LoRA adapter trained on a specific task $T_i$ is denoted as $\Delta W_i = B_i A_i$, where \( B_i \in \mathbb{R}^{d \times r} \) is the up-projection matrix, and \( A_i \in \mathbb{R}^{r \times k} \) is the down-projection matrix.
A group of LoRA task adapters is denoted by $G_j$, and its associated group adapter by $\Delta W_{G_j} = B_{G_j}A_{G_j}$. The current set of groups is written as $\mathcal{G}=\{G_1,\dots,G_M\}$, where $M$ is the current number of instantiated groups, and $G_{\max}$ denotes the maximum number of groups.

\subsection{GLAM Training Process}
\label{sec:glam_training}
\subsubsection{Task-Specific LoRA Training}
For each task $T_i$, we initialize a LoRA adapter $\Delta W_i$ and an $\alpha_i$.
The objective is for the adapter $\Delta W_i$ to learn task-relevant information that is not already captured by the base model or the previously learned groups, while the scalar $\alpha_i$ acts as a scaling factor for its contribution.
During the training phase for task $T_i$, the adapter $\Delta W_i$ and its corresponding scaling factor $\alpha_i$ are optimized using the dataset $\mathcal{D}_i$.
In this phase, the pre-trained model weights $W_0$ and the previously learned group adapters
$\{ \Delta W_{G_j} \}_{j=1}^M $ (see Sec. \ref{sec:intra_group_merge}), are kept frozen.
To dynamically change the importance of each group given the current adapter, the scaling factors of the group adapters $\{ \alpha_{G_j} \}_{j=1}^M$ are also updated while training $\Delta W_i$. This ensures that the relative importance of all adapters remains balanced when a new task is introduced. Moreover, at the classifier level, logits outside the class set of $T_i$ are masked, so that only the classifier parameters associated with the current task classes are optimized together with the task-specific representation parameters. 

Therefore, the output representation for an input \( x \) is computed as:

\begin{equation}
h_x = \underbrace{W_0 x}_\text{pre-trained model} + \underbrace{\sum_{j=1}^{M} \textcolor{red}{\alpha_{G_j}} \Delta W_{G_j} x}_\text{previous group adapters} + \underbrace{\textcolor{red}{\alpha_i \Delta W_i} x}_\text{current adapter}\label{eq:training}
\end{equation}

where the terms highlighted in red denote the components updated during task-specific training. This approach guarantees that each task-specific adapter is optimized while also accounting for the behaviors and knowledge of earlier grouped LoRAs.
By minimizing the amount of redundant knowledge across adapters, this formulation encourages the newly introduced adapter to focus solely on task-specific features, while still facilitating effective knowledge transfer from previously learned grouped adapters.

As mentioned before, GLAM adopts a dynamic grouping strategy to efficiently manage the increasing number of adapters.
This contrasts with approaches such as \citet{wu2025}, where both the number of adapters and their corresponding $\alpha$ values grow linearly with the number of tasks, leading to high computational costs.
Instead of retaining a separate LoRA module for each training experience, our strategy progressively clusters adapters by performing an initial combination phase within our sequential framework.
Under this approach, each group $G_i$ holds a single adapter $\Delta W_{G_i}$ and a unique importance factor $\alpha_{G_i}$, which greatly reduces the computational operation during training.
By consolidating adapters in this fashion, the total number of modules to be merged at inference time is substantially reduced, minimizing task interference while preserving the expressive capacity of grouped adapters.
Consequently, GLAM excels in learning across extended task sequences, mitigating forgetting while maintaining high efficiency.

\subsubsection{Adapters Grouping}
\label{sec:group_association}

\paragraph{Group Association}

After training on task $T_i$ and obtaining the corresponding adapter $\Delta W_i = B_iA_i$, GLAM assigns it to a group through a random grouping strategy. In particular, a group index is sampled uniformly from the set of available group slots up to the maximum number of groups $G_{\max}$. If the selected slot does not yet contain an adapter, a new group is initialized and $\Delta W_i$ becomes its first group adapter. Otherwise, $\Delta W_i$ is associated with the selected existing group.

Each group $G_j$ is associated with a single scaling factor $\alpha_{G_j}$, defined as the running average of the $\alpha$ values of the adapters assigned to that group. When a new adapter with importance factor $\alpha_i$ is assigned to group $G_j$, the group scaling factor is updated as

\begin{equation}
\alpha_{G_j} \leftarrow \alpha_{G_j} + \frac{\alpha_i - \alpha_{G_j}}{|G_j| + 1},\label{eq:alpha_update}
\end{equation}

where $|G_j|$ denotes the number of adapters already assigned to the group before the update. If $\Delta W_i$ initializes a new group, then $\alpha_{G_j} = \alpha_i$.

\paragraph{
Magnitude-Based Pruning}
Before merging the current adapter with the selected group, we apply 
magnitude-based pruning to retain only the 
highest-magnitude weights of the incoming adapter $\Delta W_i$. This step makes the update more selective before it is incorporated into the group representation, thereby limiting interference from weak or redundant components. 
Specifically, for matrices $B_i$ and $A_i$, we retain only the top-$k\%$ weights with the highest absolute values, where $k$ is a hyperparameter.
Hence, the resulting matrices $\hat{B}_i$ and $\hat{A}_i$ are defined as:

\begin{equation}
\begin{aligned}
    \hat{B}_i &= B_i \odot \mathbb{I}(|B_i| \geq \tau_B) \\
    \hat{A}_i &= A_i \odot \mathbb{I}(|A_i| \geq \tau_A)
\end{aligned}\label{eq:pruning}
\end{equation}

where $\odot$ represents element-wise multiplication, $\mathbb{I}(\cdot)$ is the indicator function, and $\tau_B$ and $\tau_A$ are thresholds chosen such that only the top-$k\%$ elements are retained. We prefer magnitude-based pruning over more computationally demanding importance estimates, such as Fisher-based criteria, due to its lower computational overhead and successful use in model-merging methods such as TIES~\citep{10.5555/3666122.3666432}.


\paragraph{Intra-Group Merging}
\label{sec:intra_group_merge}

To maintain a single adapter per group, we update the group representation by merging the incoming pruned adapter with the current group adapter. Specifically, let $\hat{\Delta W}_i$ be the pruned adapter associated with group $G_j$, and let $|G_j|$ denote the number of adapters already assigned to that group. The group adapter is updated through a weighted linear combination:

\begin{equation}
\Delta W_{G_j}
\leftarrow
\frac{|G_j|}{|G_j|+1}\,\Delta W_{G_j}
+
\frac{1}{|G_j|+1}\,\hat{\Delta W}_i.\label{eq:merging}
\end{equation}

Accordingly, the contribution of the previously accumulated group representation is weighted proportionally to the group size, while the incoming adapter contributes with weight $1 / (|G_j|+1)$. If a new group is created, its group adapter is simply initialized as $\Delta W_{G_j} = \Delta W_i$. 

\subsection{Classifier-alignment stage}\label{sec:ca}
After training the classifier on the current task $T_i$, the model may become biased toward predicting the most recently observed classes.
To alleviate this issue, after grouping, we freeze the feature extractor and perform a short classifier-alignment stage in which only the final linear classifier is updated. For each seen class $c$, we model the distribution of its last-layer features with a Gaussian distribution whose parameters are given by the class mean $\mu_c$ and diagonal variance $\sigma_c^2$. Because past-task data are unavailable, these statistics are computed after learning each task from its current data and stored for future alignment steps. Synthetic features are sampled as
$
\mathbf{z}_c \sim \mathcal{N}\!\left(\mu_c,\operatorname{diag}(\sigma_c^2)\right),
$
and used to retrain the classifier while keeping the representation fixed. 
Following SLCA~\citep{zhang2023slca}, we optimize the classifier using a temperature-scaled normalized cross-entropy loss:

\begin{equation}
\mathcal{L}_{\mathrm{CA}} =
\mathrm{CE}\left(
\frac{\mathbf{o}/\|\mathbf{o}\|_2}{\tau}, y
\right),
\end{equation}

where $\mathbf{o}$ denotes the classifier logits over the seen classes, $y$ is the target class label, and $\tau$ is the temperature parameter.
In our implementation, we sample 128 synthetic features per class and optimize the classifier for 5 epochs using Adam with learning rate $10^{-3}$ and temperature $\tau=0.2$.
This alignment procedure adapts the classifier to the current feature space and mitigates the bias toward newly introduced classes.

\subsection{Final Adapter Concatenation}\label{sec:final_concat}
The final stage of GLAM consists of a global concatenation step that converts the current set of group adapters into a single LoRA module, yielding a compact final parameterization of the learned groups. After the grouping phase, each group $G_j \in \mathcal{G}$ is represented by a fixed-rank adapter $\Delta W_{G_j} = B_{G_j}A_{G_j}$ and an importance factor $\alpha_{G_j}$. The concatenated adapter is obtained by scaling each group contribution by $\alpha_{G_j}$ and concatenating the resulting adapters along the rank dimension. This gives

\begin{equation}
\Delta W_{\text{final}} = \sum_{j=1}^{M} \alpha_{G_j} B_{G_j} A_{G_j}
= B_{\text{final}}A_{\text{final}}.\label{eq:concat}
\end{equation}

Since each group adapter has rank $r$, the resulting adapter has rank $r_{\text{final}} = M \cdot r$. The final model is then defined as

\begin{equation}
W_{\text{final}} = W_0 + \Delta W_{\text{final}}.
\end{equation}

This step produces a single compact adapter that summarizes the information accumulated across the current set of groups, and can be applied at any point during training.

\section{Experiments}
In this section, we present a comprehensive evaluation of the GLAM algorithm using an extensive experimental setup, providing insights into its performance and computational efficiency.

\fiftytasks


\paragraph{Experimental Setup}
We evaluate GLAM on three standard CL benchmarks: CIFAR-100, CUB-200 and ImageNet-R.
CIFAR-100 \citep{krizhevsky2009learning} includes 60,000 images over 100 classes;
CUB-200 \citep{wah2011caltech} is a fine-grained vision dataset, consisting of 11,788 bird images across 200 categories;
ImageNet-R \citep{hendrycks2021many} is a robustness benchmark consisting of 30,000 images across 200 classes, featuring artistic renditions (e.g., sketches, paintings, cartoons) of ImageNet objects to evaluate model generalization. For both CUB-200 and ImageNet-R, we use a 70\%/30\% split for training and testing, respectively.

While these datasets are widely used in the continual learning literature, evaluation protocols typically rely on a small number of sequential tasks---often around 10.
Although this constitutes a valid instantiation of the continual learning setting, it is a simplifying assumption that limits the ability to assess 
long-horizon continual learning capabilities. 
To address this, we evaluate both our proposed method and the baselines on \textbf{50-task sequences} across all benchmarks, in class-incremental learning scenarios.
This increased task horizon significantly raises the difficulty of the learning scenario, moving it closer to real-world conditions and providing a more rigorous assessment of each method’s scalability.

We report two standard CL metrics: (i) \textbf{Average Accuracy (AA)}: mean accuracy over all tasks at the end of training; (ii) \textbf{Forgetting Measure (FM)}~\citep{lopez2017gradient}: 
the average drop between the accuracy measured immediately after learning a task and its final accuracy after training on all tasks.
Alongside these metrics, we also perform an efficiency analysis focusing on two key aspects: time complexity, measured by training time, and space complexity, measured by the number of trainable parameters introduced by each method.

For the backbone model, we used a ViT-B/16 backbone pretrained on ImageNet.
For GLAM, we apply pruning to retain the top 80\% of weights per adapter, and use \(G_\text{max}=5\) task groups; additional hyperparameter details are provided in Appendix~\ref{sec:glam_hparams}. Training is done using the Adam optimizer \citep{loshchilov2017decoupled}, with a learning rate of \(10^{-3}\) and batch size of 64.
We compare against state-of-the-art parameter-efficient continual learning methods, considering both prompt-based and LoRA-based methods as PEFT baselines.
For L2P and CODA-Prompt we used the implementation available in the \texttt{mammoth} library \citep{boschini2022class}.
For SEMA, InfLoRA, SD-LoRA and CL-LoRA we used the official implementation from the authors.
Since our method involves a merging phases, we also benchmark it against the widely used Linear Merging technique.
In this setting, a task-specific adapter is trained for each task and then merged with the adapter representing the previously learned tasks, using the implementation from the Hugging Face \texttt{peft} library \citep{peft}.
For all baselines, we adopt the default method-specific hyperparameters from their original configurations and follow the corresponding published protocols.
Crucially, evaluating efficiency metrics in a meaningful way requires a uniform parameter budget; we therefore fix the LoRA rank to $r=16$ across GLAM and all LoRA-based baselines.
We do not apply classifier alignment to the baselines in
Table~\ref{tab:50tasks}, as this would modify their original configurations and computational budgets.  For CL-LoRA, such alignment is not directly applicable because inference relies on class prototypes rather than the learned linear classifier. Moreover, for InfLoRA and SEMA, the original studies report only marginal gains from related bias-mitigation techniques despite their additional computational cost, and therefore the authors use the unaligned variants as the default configurations.


\subsection{Results Analysis}
In Table \ref{tab:50tasks}, we report accuracy and forgetting across the selected benchmarks under a challenging 50-task training sequence.
GLAM significantly outperforms state-of-the-art continual learning methods in this setting, where task sequences are substantially longer than in standard evaluations.
GLAM achieves the highest accuracy on all three benchmarks, consistently surpassing competing approaches.
In contrast, several methods exhibit strong performance on specific datasets but fail to generalize across scenarios.
For instance, SEMA achieves comparable results on ImageNet-R but trails GLAM by approximately 20 percentage points on CUB-200.
Overall, GLAM demonstrates superior and more consistent performance across all evaluation settings, demonstrating its efficacy on longer, more complex learning scenarios.\\
In conjunction with the accuracy results, we also report the evaluation of the forgetting metric.
Compared to the strongest baselines in terms of accuracy, namely SEMA, SD-LoRA and CL-LoRA, GLAM achieves comparable forgetting performance.
These results, together with the accuracy gains, highlight the effectiveness of our grouping strategy in mitigating interference during the merging phase, thereby improving both knowledge retention and predictive performance.

\begin{figure}
    \centering
    \begin{subfigure}{0.33\textwidth}
        \centering
        \includegraphics[width=\linewidth]{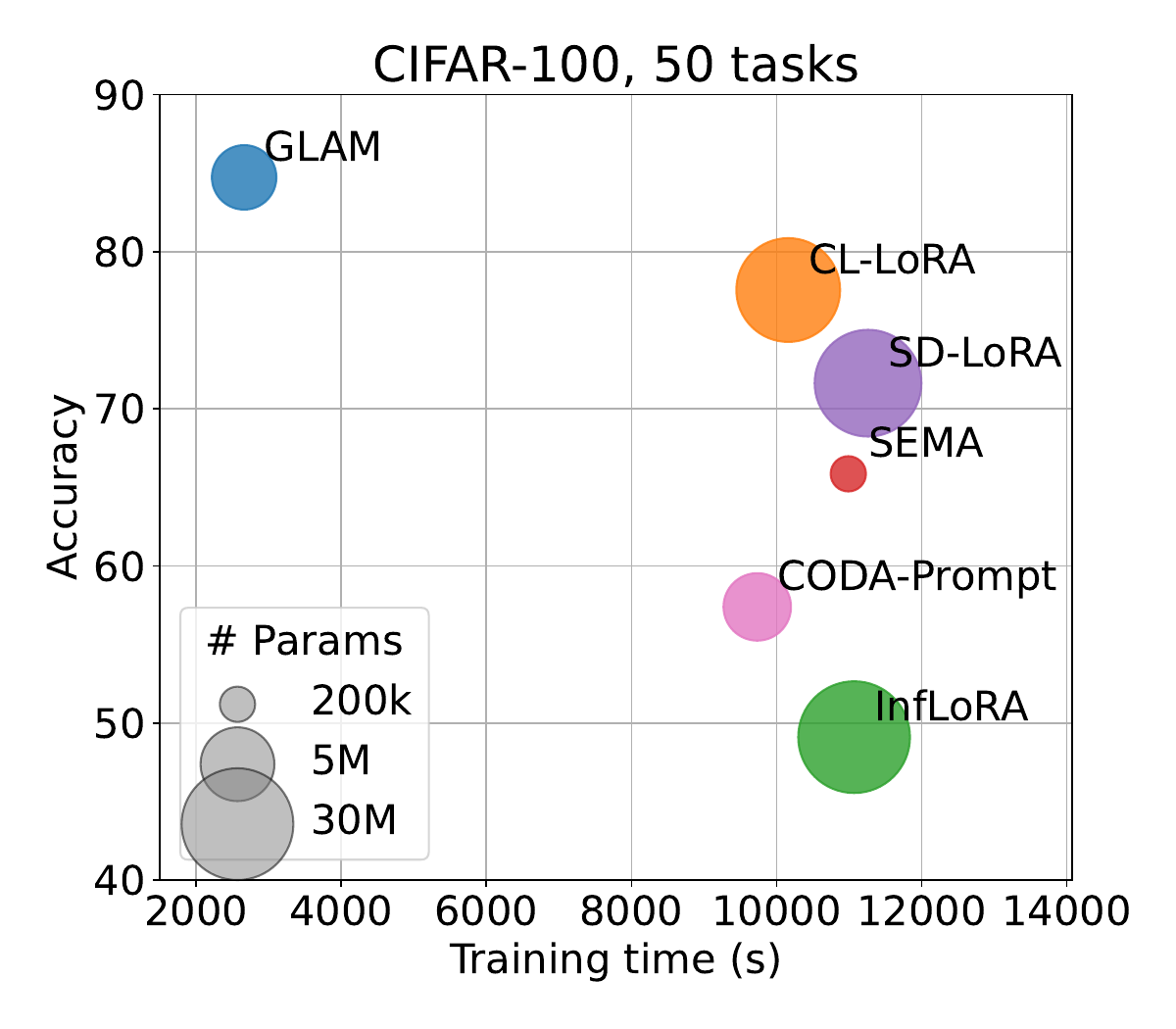}
    \end{subfigure}
    \hfill
    \begin{subfigure}{0.33\textwidth}
        \centering
        \includegraphics[width=\linewidth]{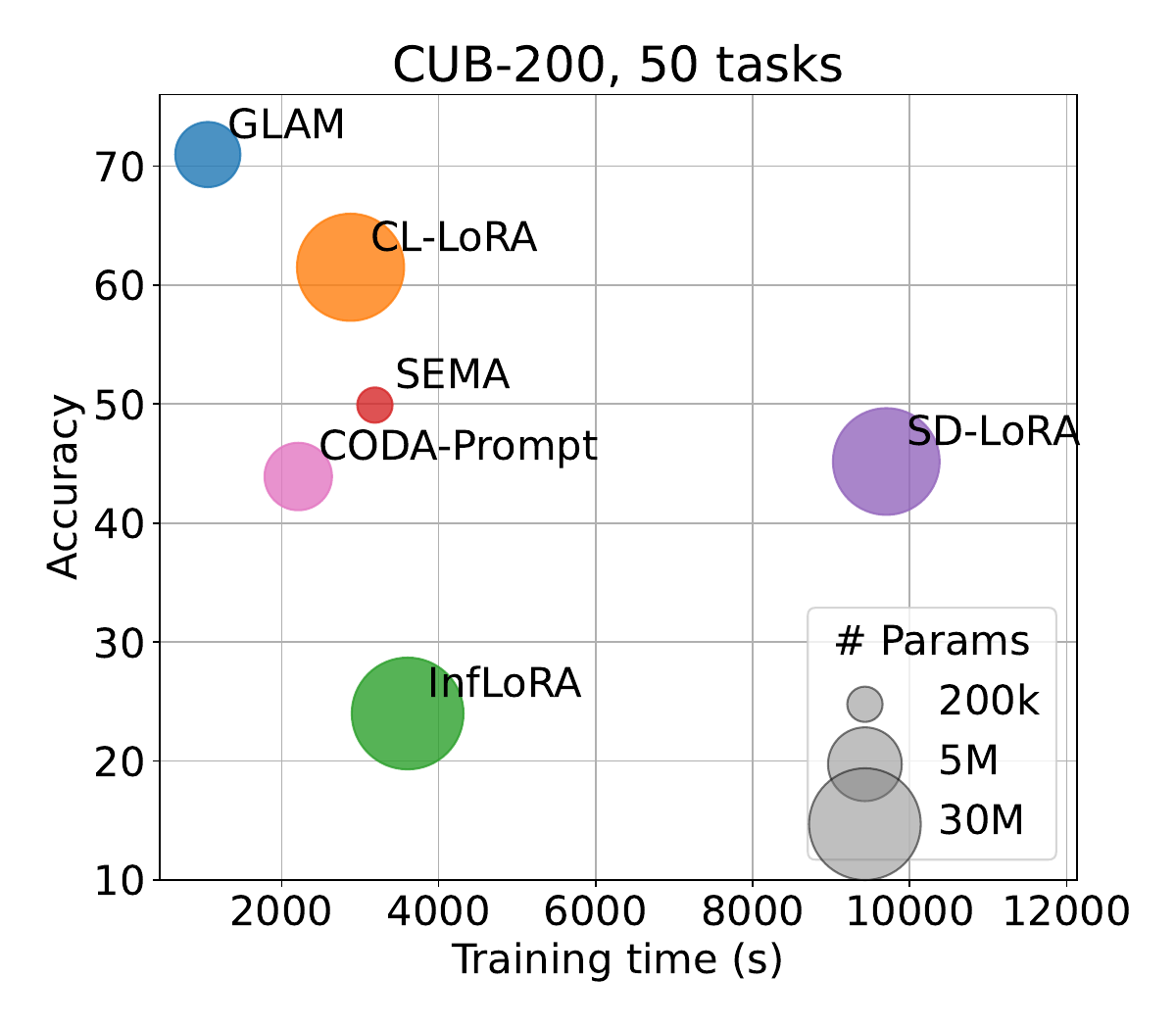}
    \end{subfigure}
    \hfill
    \begin{subfigure}{0.33\textwidth}
        \centering
        \includegraphics[width=\linewidth]{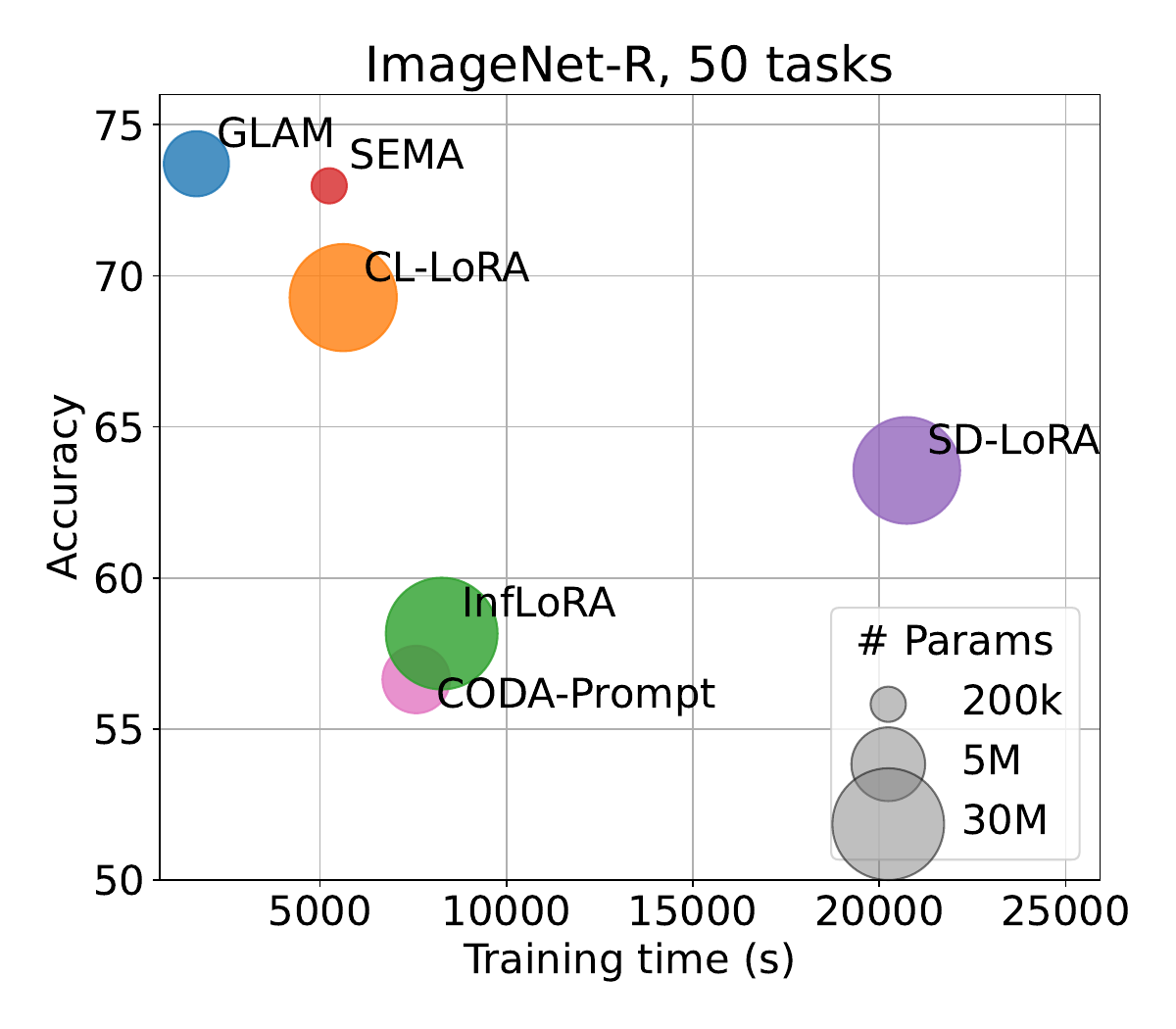}
    \end{subfigure}
    \caption{Efficiency--performance trade-off over the 50-task setting across all benchmarks. GLAM consistently occupies the most favorable region of the plot, combining higher accuracy with lower training time and competitive parameter usage relative to prior methods.}
    \label{fig:efficiency_all}
\end{figure}

\paragraph{Efficiency Analysis}
In addition to accuracy, Table \ref{tab:50tasks} also reports 
computational efficiency in terms of wall-clock training time, measured for the complete training procedure of each method. For GLAM, this includes task-adapter training, pruning, merging, computation of class statistics, and the subsequent classifier-alignment stage.
This trade-off is further illustrated in Figure~\ref{fig:efficiency_all}, where the horizontal axis reports training time, the vertical axis reports final average accuracy, and the size of each marker is proportional to the number of trainable parameters. As shown in the figure, GLAM consistently lies in the most favorable region of the plot, achieving the highest accuracy while requiring only a fraction of the training time and competitive parameter usage compared to prior methods. In particular, GLAM 
is approximately $3\times$ faster than the strongest-performing baseline (CL-LoRA), while using about $5\times$ fewer trainable parameters. This efficiency gain is largely due to GLAM's grouped-merging strategy and pruning approach, which reduce the number of active parameters and lower computational overhead during training and inference.
These results highlight GLAM’s scalability to real-world scenarios, where time and computational resources are constrained over longer task sequences.

\paragraph{Performance Across Different Sequence Lengths}

To evaluate the effect of sequence length, we report results for 10-, 20-, and 50-task settings in Figure~\ref{fig:sequence_lenght}, and further extend the analysis to 100-task sequences in Appendix~\ref{sec:100_tasks}. Each sequence length corresponds to a separate continual-learning experiment with a different task partition.
Across all three benchmarks, GLAM shows the most favorable overall trend as the number of tasks grows. On CIFAR-100, its accuracy remains nearly unchanged from 20 to 50 tasks, while the competing methods decline more noticeably. On CUB-200 and ImageNet-R, where longer sequences lead to a larger drop for all methods, GLAM still degrades substantially less than the baselines.
These results underscore the effectiveness of the GLAM framework, which remains robust as the number of tasks increases and the continual learning problem becomes more challenging, while competing baselines exhibit a noticeable performance degradation.
\begin{figure}
    \centering
    \begin{subfigure}{0.33\textwidth}
        \centering
        \includegraphics[width=\linewidth]{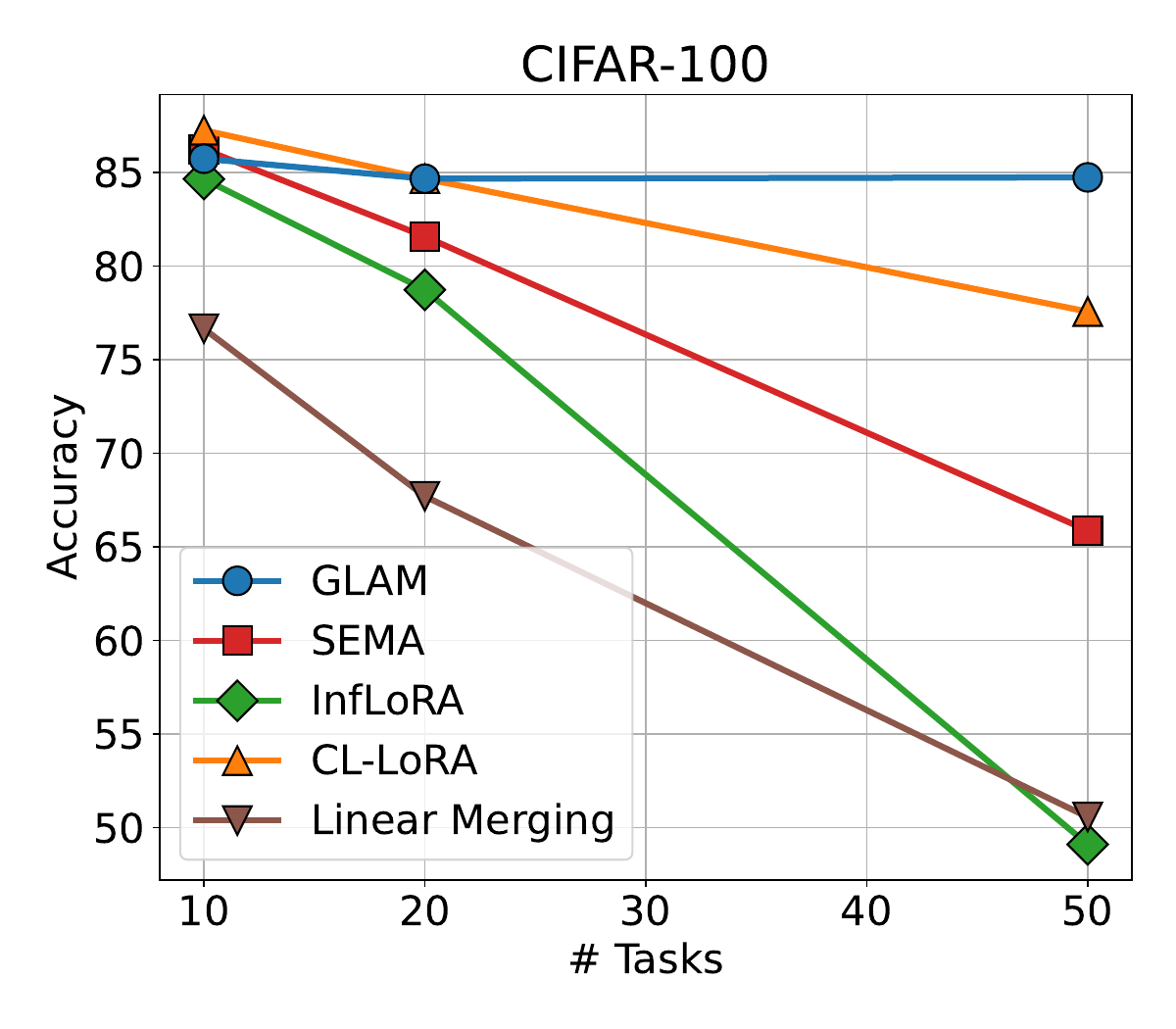}
    \end{subfigure}
    \hfill
    \begin{subfigure}{0.33\textwidth}
        \centering
        \includegraphics[width=\linewidth]{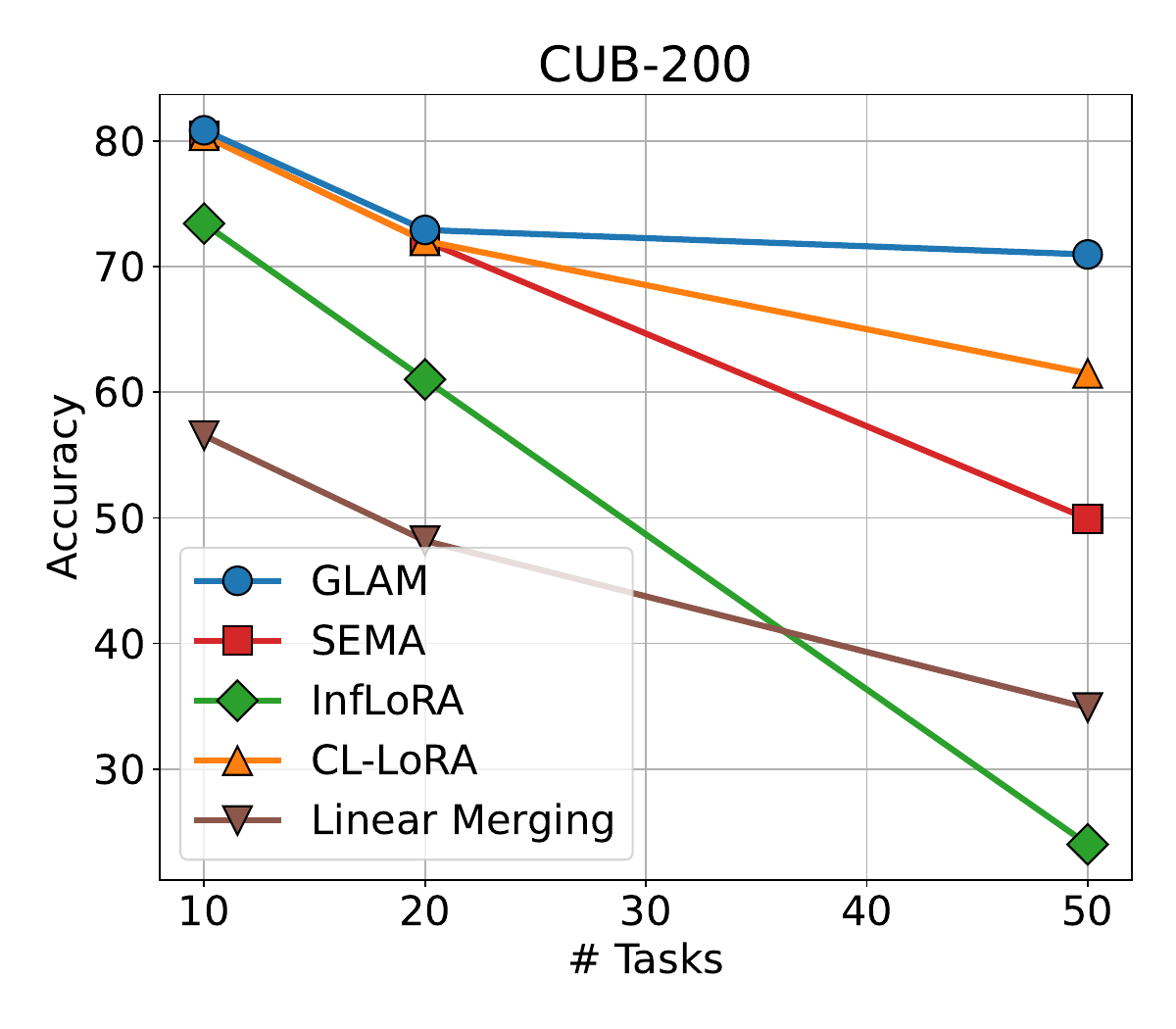}
    \end{subfigure}
    \hfill
    \begin{subfigure}{0.33\textwidth}
        \centering
        \includegraphics[width=\linewidth]{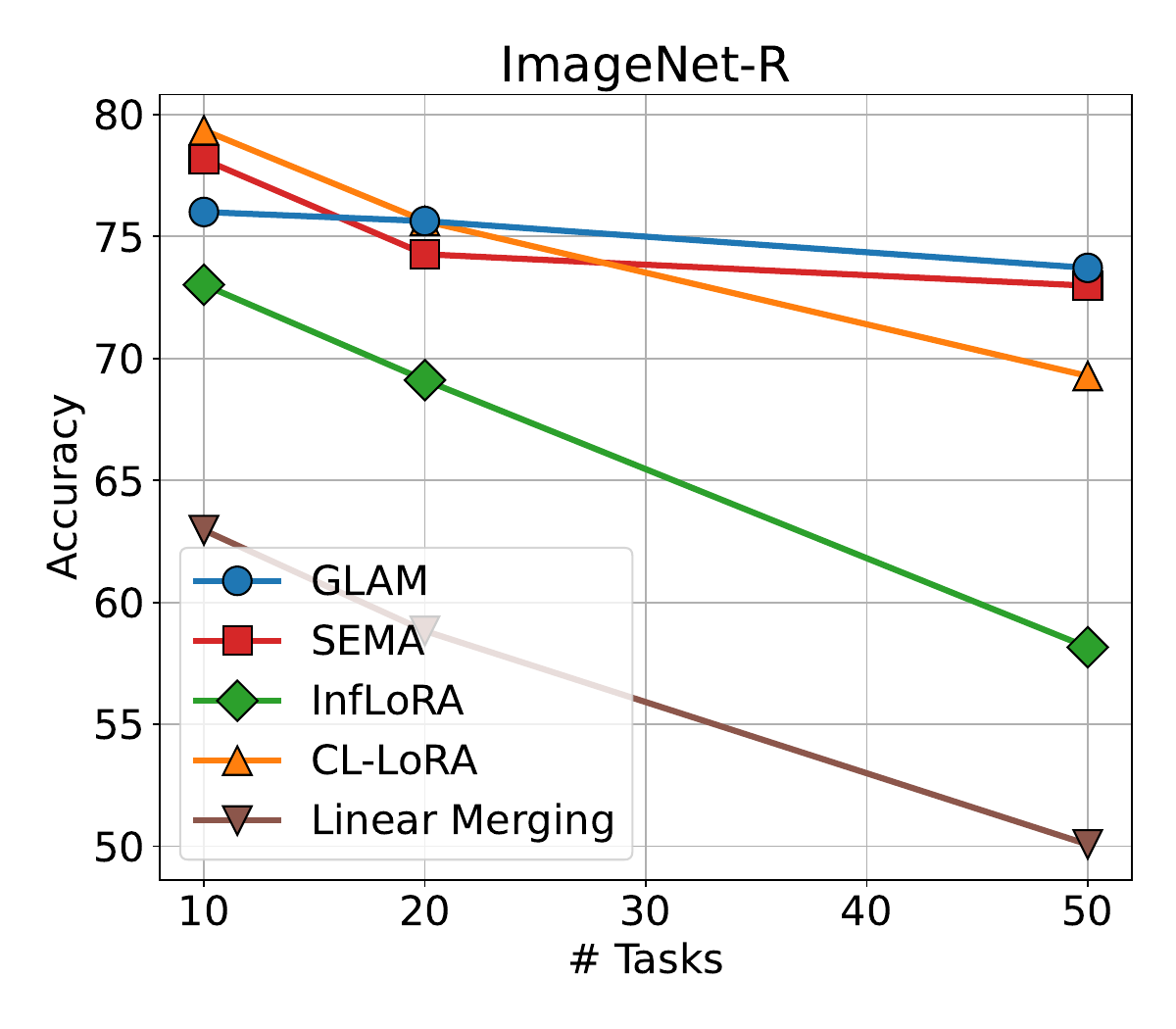}
    \end{subfigure}
    \caption{Final average accuracy of the compared methods on CIFAR-100, CUB-200, and ImageNet-R with 10, 20 and 50 tasks. As the number of tasks increases, GLAM consistently outperforms the SOTA baseline across all benchmarks.}
    \label{fig:sequence_lenght}
\end{figure}

\AblSimGroupImpact

\subsection{Ablation Studies}
\label{sec:abl}

\paragraph{Impact of the Grouping Algorithms}
One of the key components of GLAM is its grouping strategy, which merges multiple adapters into a single module.
Table \ref{tab:similarity_orthogonal} reports the performance of three different grouping algorithms (see Appendix \ref{app:alternative_grouping}).
The results show no clear winner, as all strategies achieve comparable performance across benchmarks on average.
Given this, and to prioritize computational efficiency, we adopt random grouping as part of our main methodology.

The limited impact of the grouping strategy can be explained by two main factors.
First, the pruning and merging stages act as a strong form of regularization, suppressing redundant or interfering components and thereby mitigating suboptimal grouping decisions.
Second, during task-specific training, predictions depend on both the incoming adapter and the previously learned group adapters (Eq.~\ref{eq:training}). This encourages the new adapter to capture information complementary to the existing groups, reducing sensitivity to its subsequent group assignment.
As a result, more sophisticated criteria, such as similarity or orthogonality, do not offer a significant advantage over simpler approaches, such as random grouping, which achieve comparable performance at a lower computational cost.

\paragraph{Impact of the Classifier Alignment Stage}
Table \ref{tab:classifier_alignment} provides insights into the impact of the classifier alignment stage on the overall performance of GLAM.
We compare the full model with a variant without classifier alignment, and additionally evaluate a simple merging baseline, both with and without the alignment procedure. These results highlight two key findings.
First, classifier alignment proves to be highly effective, significantly improving the performance of both GLAM and the baseline on ImageNet-R with 50 tasks.
However, even with this enhancement, the baseline remains markedly inferior to GLAM, indicating that the proposed grouping mechanism is essential for achieving strong performance over long task sequences.

\AblGroupingMergingPruning

\paragraph{Impact of the Number of Groups}
Table \ref{tab:impact_grouping_performance} 
shows that GLAM maintains strong performance even with a small number of groups.
In particular, increasing the number of groups from the extreme case of a single group to ten yields an improvement of 3\% in accuracy, while further increases result in diminishing returns.

This suggests that constructing larger final adapters---obtained by aggregating more group adapters---may introduce overparameterization and redundancy.
Nevertheless, GLAM remains robust across different grouping configurations, exhibiting only modest performance variations.
For this reason, to balance accuracy and computational efficiency, we adopted five groups in our main experiments while still achieving top performance. Importantly, $G_{\max}$ need not scale proportionally with the number of tasks, which may be unknown in advance. The 100-task results in Appendix~\ref{sec:100_tasks}, obtained with the same value $G_{\max}=5$, further show that this fixed configuration remains effective for longer task sequences.


\paragraph{Impact of Pruning}
Table \ref{tab:ablation_pruning} shows GLAM's performance when varying the percentage of retained weights ($k$).
The average accuracy 
improves as $k$ increases from 20\% to 80\%, then slightly declines at $k=100\%$.
This suggests that pruning effectively balances information retention and noise reduction.
When $k$ is small, aggressive pruning restricts the amount of new task-specific information incorporated into the group adapter; as a result, forgetting is reduced, but learning is also weaker. As $k$ increases, the model can absorb richer task-specific updates, improving average accuracy up to an intermediate regime. However, retaining all weights may also preserve redundant or noisy components, which can slightly hurt performance.

\paragraph{Impact of the Merging Algorithm}
A crucial step in GLAM is the intra-group merging stage, where each newly trained LoRA module is merged into a selected group adapter.
To assess its impact, we compare the two main merging strategies on a 50-task ImageNet-R scenario, while also varying the percentage of retained weights (Table \ref{tab:merging_algorithms}).
Interestingly, the simpler linear merging strategy consistently outperforms the more sophisticated TIES method across all pruning levels.
This behavior is likely due to the asymmetric nature of the intra-group merge phase, where a newly pruned adapter is combined with an existing group adapter that is both denser and assigned a larger importance weight. In this regime, the sign-resolution mechanism of TIES may be biased toward the group adapter and attenuate the contribution of the new task update, whereas linear merging preserves both updates more directly.

\section{Conclusion}
This work introduces GLAM, a novel approach for class-incremental learning based on a grouped LoRA merging mechanism.
GLAM follows a two-stage procedure: first, adapters are merged within groups; then, they are rescaled using group-specific importance factors and combined through group-wise concatenation.
We place particular emphasis on a more realistic evaluation setting.
On one hand, we consider longer task sequences than those typically used in standard protocols, enabling a more faithful assessment of lifelong learning capabilities.
On the other hand, we explicitly focus on computational efficiency, carefully designing each component of the method and selecting hyperparameters to balance effectiveness and efficiency.
Through extensive experiments, we show that GLAM consistently outperforms state-of-the-art baselines in terms of accuracy, while requiring only a fraction of the training time and number of trainable parameters compared to the strongest-performing baseline.
These results demonstrate the effectiveness of the proposed pipeline in mitigating interference among adapters at scale, while maintaining a lightweight and efficient framework compared to competing approaches.

\section*{Acknowledgments}
This research was partially supported by the FIS2 Grant from the Italian Ministry of University
and Research (MUR), Grant No. FIS2023-03382, under the project ``Continual, Decentralized Compositionality for
Sustainable Artificial Intelligence'' with Prof. Vincenzo Lomonaco serving as Principal Investigator (PI).


\bibliography{references}
\bibliographystyle{collas2026_conference}

\clearpage
\appendix
\section{Appendix}
\subsection{Computational Resources}

All experiments were conducted on the Leonardo supercomputer at CINECA. Table~\ref{tab:system} provides detailed specifications of the computational environment used for reproducibility.

\begin{table}[h]
\centering
\renewcommand{\arraystretch}{1.2}
\begin{tabular}{p{3.5cm} | p{10cm}}
\toprule
\textbf{System} & \textbf{Details} \\
\midrule
\textbf{Cluster} & Leonardo @ CINECA \\
\textbf{OS} & Red Hat Enterprise Linux 8.7 (Ootpa) \\
\midrule
\textbf{Booster Module (GPU)} & Atos BullSequana X2135 "Da Vinci" blades \\
\textbf{Booster Nodes} & 3456 compute nodes \\
\textbf{CPU} & 32 × Intel Ice Lake @ 2.60 GHz \\
\textbf{GPU} & 4 × NVIDIA A100 (Ampere), 64 GB \\
\textbf{RAM} & 512 GB per node \\
\midrule
\textbf{DCGP Module (CPU)} & Atos BullSequana X2140 blades \\
\textbf{DCGP Nodes} & 1536 compute nodes \\
\textbf{CPU} & 2 × 56-core Intel Sapphire Rapids @ 2.00 GHz \\
\textbf{RAM} & 512 GB per node \\
\midrule
\textbf{Network} & 200G HDR Infiniband Dragonfly+ \\
\bottomrule
\end{tabular}
\caption{System configuration and environment used for all experiments.}
\label{tab:system}
\end{table}

\subsection{GLAM Algorithm}
\label{app:algorithm}
In this section, we detail the GLAM algorithm through pseudo-code presented in Algorithm \ref{alg:GLAM}, offering a clear and reproducible outline of its computational steps.
\Algo


\subsection{GLAM Hyperparameters}\label{sec:glam_hparams}
Table~\ref{tab:glam_hparams} reports the main hyperparameters used in GLAM.
Table~\ref{tab:lora_rank} reports the performance of the considered LoRA-based methods for $r \in \{8,10,16\}$, with all other hyperparameters kept unchanged. GLAM achieves the highest accuracy for every tested rank on both ImageNet-R and CUB-200. Moreover, its strongest results are obtained with $r=8$, exceeding those reported in Table~\ref{tab:50tasks} with $r=16$. Overall, the relative performance of the methods remains consistent across the evaluated ranks.

\begin{table}
\centering
\begin{tabular}{lc}
\toprule
\textbf{Parameter} & \textbf{Value} \\
\midrule
\multicolumn{2}{c}{LoRA Parameters} \\
\midrule
Rank & 16 \\
$\alpha$ & 16 \\
Target Modules & QKV projection \\
Dropout & 0.1 \\
Bias & None \\
\midrule
\multicolumn{2}{c}{Optimization Parameters} \\
\midrule
Optimizer & Adam \\
Weight decay & 0.0 \\
Epochs per task & 10 \\
Learning rate  & 1e-3 \\
\midrule
\multicolumn{2}{c}{Grouping Parameters} \\
\midrule
Number of groups & 5 \\
$k$ & 0.8 \\
Grouping mechanism & Random \\
Merging technique & Linear \\
\midrule
\multicolumn{2}{c}{Classifier Alignment Stage Parameters} \\
\midrule
Synthetic samples per class & 128 \\
Epochs per stage & 5 \\
Learning rate  & 1e-3 \\
Temperature & 0.2 \\
\bottomrule
\end{tabular}
\caption{Hyperparameters used in GLAM across the main experiments.}
\label{tab:glam_hparams}
\end{table}

\begin{table}[t]
\centering
\small
\begin{tabular}{llcccc}
\toprule
\textbf{Benchmark}
& \textbf{Rank}
& \textbf{InfLoRA}
& \textbf{SD-LoRA}
& \textbf{CL-LoRA}
& \textbf{GLAM} \\
\midrule

\multirow{3}{*}{ImageNet-R}
& 8 & 55.70 $\pm$ 0.29 & 71.51 $\pm$ 0.86 & 69.44 $\pm$ 0.04 & \textbf{74.00 $\pm$ 0.16} \\

& 10 & 55.70 $\pm$ 0.71 & 70.12 $\pm$ 0.70 &  69.92 $\pm$ 0.55 & \textbf{73.66 $\pm$ 0.33}\\

& 16
& 58.16 $\pm$ 0.69
& 63.56 $\pm$ 1.76 
& 69.28 $\pm$ 1.06
& \textbf{73.71 $\pm$ 0.68} \\

\midrule

\multirow{3}{*}{CUB-200}
\multirow{3}{*}{CUB-200}
& 8 & 22.85 $\pm$ 2.97 & 40.76 $\pm$ 2.11 & 61.23 $\pm$ 1.15 & \textbf{74.86 $\pm$ 1.15} \\

& 10 & 20.52 $\pm$ 4.94 & 41.95 $\pm$ 2.55 & 62.41 $\pm$ 2.77 & \textbf{73.89 $\pm$ 1.56} \\

& 16 & 24.00 $\pm$ 2.71 & 45.17 $\pm$ 1.44 & 61.49 $\pm$ 1.90 & \textbf{70.96 $\pm$ 2.06} \\


\bottomrule
\end{tabular}
\caption{Accuracy (\%) of different LoRA-based methods across varying LoRA ranks on
ImageNet-R and CUB-200 under the 50-task setting.
}
\label{tab:lora_rank}
\end{table}

\subsection{Additional Baselines}

Table~\ref{tab:add_baselines} reports additional baseline comparisons, including joint training approaches where the model learns all tasks simultaneously. Joint training serves as an upper bound for performance, as it has access to all task data concurrently.
In particular, \textit{joint} refers to the full fine-tuning of the model on all data, while \textit{joint LoRA} indicates the performance when training a single LoRA adapter on all tasks. In contrast, \textit{naive} denotes sequential full fine-tuning, in which the model is trained on each task in turn using only the current task data.

\begin{table}[h]
    \centering
    \renewcommand{\arraystretch}{1.2}
    \begin{tabular}{p{4cm} | p{3cm} p{3cm}}
        \hline
        \textbf{Method} & \textbf{CIFAR-100 (AA)} & \textbf{CUB-200 (AA)} \\
        \hline
        Joint        & 83.11 ± 0.50 & 78.56 ± 2.17 \\
        Joint LoRA   & 86.58 ± 1.27 & 37.53 ± 1.65 \\
        \hline
        Naive & 50.27 ± 3.97 & 11.97 ± 1.82 \\
        \hline
        SEMA         & 65.86 ± 0.34 & 49.91 ± 2.60 \\
        InfLoRA      & 49.10 ± 4.40 & 24.00 ± 2,72 \\
        SD-LoRA      & 71.63 ± 0.67 & 45.17 ± 1.44 \\
        CL-LoRA      & 77.56 ± 1.43 & 61.49 ± 1.90 \\
        \hline
        \textbf{GLAM} & \textbf{84.73 ± 0.74} & \textbf{70.96 ± 2.06} \\
        \hline
    \end{tabular}
    \caption{Average Accuracy (\%, $\uparrow$) on CIFAR-100 and CUB-200 across 50 tasks. Joint performance represents an upper bound.}
    \label{tab:add_baselines}
\end{table}

\subsection{Additional Results on 100-Task Sequences}\label{sec:100_tasks}
To further assess the behavior of GLAM over longer task horizons, we evaluate both GLAM and CL-LoRA, the strongest competing baseline in our main experiments, on 100-task partitions of CUB-200 and ImageNet-R. The results are reported in table~\ref{tab:acc_100tasks}. GLAM maintains strong performance in the 100-task setting relative to the corresponding 50-task setting. On CUB-200, accuracy remains nearly unchanged, while on ImageNet-R GLAM retains high accuracy and shows a much smaller degradation than CL-LoRA. These results provide additional evidence that GLAM scales favorably to longer task sequences.
\begin{table}
\centering
\begin{tabular}{llcc}
\toprule
\textbf{Dataset} & \textbf{Number of tasks} & \textbf{CL-LoRA} & \textbf{GLAM} \\
\midrule
CUB-200 & 50 & 61.49 $\pm$ 1.90 & \textbf{70.96} $\pm$ 2.06 \\
CUB-200 & 100 & 65.77 $\pm$ 1.00 & \textbf{70.78} $\pm$ 0.52 \\
ImageNet-R & 50 & 69.28 $\pm$  1.06 & \textbf{73.71} $\pm$ 0.68 \\
ImageNet-R & 100 & 52.6 $\pm$ 5.22 & \textbf{72.15} $\pm$ 0.65 \\
\bottomrule
\end{tabular}
\caption{Average Accuracy (\%, $\uparrow$) on CUB-200 and ImagetNet-R across 50 and 100 tasks.}
\label{tab:acc_100tasks}
\end{table}

\subsection{Robustness to Noisy Labels}\label{app:noisy_labels}
To evaluate the robustness of GLAM under imperfect supervision, we conduct additional experiments with synthetic label noise on CUB-200 with 50 tasks. Specifically, a fraction of the training samples is randomly assigned an incorrect label, while evaluation is performed on the original clean test set. We compare GLAM against CL-LoRA, the strongest competing baseline in our main experiments. Table~\ref{tab:noisy_labels} reports the resulting average accuracy. GLAM maintains strong performance across all noise levels and consistently outperforms CL-LoRA. While both methods experience some degradation at higher noise rates, the results indicate that the proposed grouping and merging strategy remains effective even when the training signal is substantially corrupted.

\begin{table}
\centering
\begin{tabular}{lcc}
\toprule
\textbf{Noisy labels (\%)} & \textbf{CL-LoRA} & \textbf{GLAM} \\
\midrule
0 & 61.49 & \textbf{70.96} \\
20 & 63.65 & \textbf{71.58} \\
40 & 59.73 & \textbf{67.00} \\
\bottomrule
\end{tabular}
\caption{Average Accuracy (\%, $\uparrow$) under different levels of synthetic label noise.}
\label{tab:noisy_labels}
\end{table}

\subsection{Details on Alternative Grouping Criteria} \label{app:alternative_grouping}
In addition to the random grouping strategy adopted in the main method, we also considered similarity-based and orthogonality-based criteria for assigning a task adapter to an existing group.

After training on task $T_i$, let $\Delta W_i = B_iA_i$ denote the corresponding adapter. To compare $\Delta W_i$ with a previously learned group adapter $\Delta W_{G_j}$, we compute a score based on the last LoRA layer $L$ of the two adapters:
\begin{equation}
S(\Delta W_i,\Delta W_{G_j})
=
\frac{
\cos(A_i^L, A_{G_j}^L)
+
\cos(B_i^L, B_{G_j}^L)
}{2},
\qquad j \in \{1,\dots,M\},
\end{equation}
where $\cos(\cdot,\cdot)$ denotes cosine similarity after vectorizing the corresponding LoRA factors.

In the \emph{similarity-based} variant, $\Delta W_i$ is assigned to the group with highest score, provided that the score exceeds a threshold $\tau_{\mathrm{sim}}$. If no group satisfies this condition and the maximum number of groups has not yet been reached, i.e., $M < G_{\max}$, a new group is created. Otherwise, if $M = G_{\max}$, the adapter is assigned to the group with highest similarity regardless of the threshold.

In the \emph{orthogonality-based} variant, the same score is used, but lower values are preferred, with the goal of assigning the incoming adapter to the most dissimilar group. In this case, $\Delta W_i$ is assigned to the group with lowest score, provided that the score is below a predefined threshold. As above, if no group satisfies the threshold and $M < G_{\max}$, a new group is created; otherwise, when $M = G_{\max}$, the adapter is assigned to the group with lowest score.


\subsection{Declaration on Generative AI}
  
 During the preparation of this work, the authors used OpenAI ChatGPT for grammar and spelling check, paraphrase and reword. After using this tool, the authors reviewed and edited the content as needed and take full responsibility for the publication’s content.

\end{document}

%% file: math_commands.tex
\usepackage{amsmath,amsfonts,bm}

\def\eqref#1{equation~\ref{#1}}

\def\1{\bm{1}}

\DeclareMathAlphabet{\mathsfit}{\encodingdefault}{\sfdefault}{m}{sl}
\SetMathAlphabet{\mathsfit}{bold}{\encodingdefault}{\sfdefault}{bx}{n}

